\documentclass[11pt]{article}

\usepackage[preprint]{acl}

\usepackage{times}
\usepackage{latexsym}
\usepackage{amsmath}
\usepackage{amssymb}
  \usepackage{xcolor}
  \newcommand{\best}[1]{\textcolor{blue!70!black}{\textbf{#1}}}
  \newcommand{\second}[1]{\textcolor{blue!55!black}{#1}}
  \newcommand{\bad}[1]{\textcolor{red!70!black}{#1}}
\usepackage[T1]{fontenc}

\usepackage[utf8]{inputenc}

\usepackage{microtype}

\usepackage{inconsolata}
\usepackage{xurl}

\usepackage{graphicx}
\usepackage{booktabs}
\usepackage{array}
\usepackage{tabularx}
\usepackage{placeins}
\usepackage{float}
\usepackage{tikz}
\usepackage{makecell}
\usepackage{pifont}
\usetikzlibrary{arrows.meta,positioning,shapes.geometric,fit}
\newcommand{\cmark}{\ding{51}}
\newcommand{\xmark}{\ding{55}}

\title{FraudSMSWalker: Benchmarking Agentic Large Language Models for SMS-to-Webpage Fraud Detection}

\author{
\begin{tabular}{c}
Yanhan Zhou\textsuperscript{*} \quad
Zhiming Ma\textsuperscript{*} \quad
Yijin Zhou\textsuperscript{*} \quad
Yutong Li\textsuperscript{*} \quad
Hanxiao Xiang \\
Yiming Cheng \quad
Tailai Chen \quad
Kaijie Zhang \quad
Zhihao Nan \quad
Jiahao Ni \\
Zhao Wu\textsuperscript{\textdagger} \quad
Qingyun Pan \quad
Shun Zhang\textsuperscript{\textdagger} \quad
Shi Cheng \quad
Mingyu Luo
\end{tabular}
}

\begin{document}
\maketitle
\begingroup
\renewcommand{\thefootnote}{}
\footnotetext{\textsuperscript{*}These authors contributed equally to this work.\\
\textsuperscript{\textdagger}Corresponding authors: zhaowu@baimaohui.net, shunzhang@ppsuc.edu.cn}
\endgroup
\begin{abstract}
SMS fraud is increasingly cross-channel: a message directs the user to a webpage, and the final risk depends on how the SMS claim aligns with the page content and requested user action. However, existing evaluations either focus on message-only smishing classification or expose URL and domain cues that allow models to rely on reputation shortcuts. To address this gap, we introduce \textbf{FraudSMSWalker}, a controlled benchmark for URL-masked SMS-to-webpage fraud judgment. FraudSMSWalker contains 699 bilingual chains, including 332 fraudulent and 367 benign cases, across ten service scenarios. The model-visible input consists of the SMS context and sanitized webpage evidence, while raw URLs, hosts, domains, IPs, redirects, and reputation metadata are withheld. The benchmark further includes hard benign cases whose pages contain login, payment, verification, or account-management elements that are plausible under the service context but also appear in scam flows. We evaluate nine web agents under masked browser-agent protocols and conduct URL-visibility ablations. The results show that current agents can detect suspicious cues, but struggle to preserve benign recall and often produce positive predictions that are weakly supported by the observed evidence. These findings position FraudSMSWalker as a benchmark for measuring whether web agents can make fraud judgments that remain both accurate and evidence-grounded when direct reputation shortcuts are suppressed. The associated code and dataset are accessible at the \href{https://anonymous.4open.science/w/FraudMessageWalker-Bench}{anonymous link}.
\end{abstract}

\section{Introduction}
\label{sec:introduction}

SMS scams increasingly unfold as cross-channel interactions. A message may claim that a parcel is delayed, a payment failed, or an account requires verification, then direct the user to a webpage that resembles a routine service portal. In such cases, the risk is not determined by the SMS or the webpage in isolation. It depends on whether the message claim, page content, and requested user action jointly form a deceptive service flow. We refer to this setting as \emph{SMS-to-webpage fraud judgment}.

This setting is practically important and difficult to evaluate. The U.S. Federal Trade Commission reports \$12.5 billion in consumer fraud losses in 2024, including \$470 million from text-message scams \citep{ftc2025fraudloss,ftc2025textscams}; the FBI IC3 reports over \$16.6 billion in internet-crime losses for the same year \citep{fbi2025ic3}. Operational defenses often rely on URL blocklists, domain reputation, redirects, search results, or third-party scanners \citep{mahmood2023smishing,altay2024phishingurl,xiang2013phishingwebpage}. These signals are useful in deployed systems, but they complicate model evaluation: a high score may reflect the use of location or reputation cues rather than evidence from the SMS and webpage content.

Existing benchmarks leave this issue unresolved. Text-only smishing datasets test whether a model can classify a message, but they omit the webpage evidence that often determines the user-facing risk. Phishing and web-agent benchmarks include webpages, but frequently expose URLs or domains that can act as shortcuts. General web-agent benchmarks further emphasize navigation, retrieval, and task completion \citep{zhou2024webarena,lu2024weblinx}, rather than safety-critical judgment under controlled evidence. As a result, current evaluations do not isolate whether agents can make fraud judgments from the visible SMS-to-webpage chain when direct location and reputation cues are hidden.

We introduce \textbf{FraudSMSWalker}, a controlled benchmark for URL-masked SMS-to-webpage fraud judgment. It contains 699 bilingual SMS-to-webpage chains, with 332 fraudulent and 367 benign cases, across ten common service scenarios. Each case pairs an SMS-side claim anchored by a real-world SMS message or report with sanitized webpage evidence, while raw URLs, hosts, domains, IPs, redirects, and reputation metadata are withheld from model-visible inputs. The benchmark is designed to make false-positive control measurable: many benign cases contain login forms, payment widgets, verification steps, or account-management elements that are legitimate within the stated service flow, even though similar interface patterns also appear in fraudulent flows.

We evaluate nine web agents under masked browser-agent protocols and conduct URL-visibility ablations. The main finding is that current agents often identify suspicious cues but fail to preserve benign recall when direct location cues are hidden. This leads to many false positives on legitimate service flows. The evidence-support audit further shows that many positive predictions are not well grounded in the observed trajectory. These results suggest that the main bottleneck is not only fraud sensitivity, but the ability to make restrained and evidence-supported judgments.

Our contributions are threefold. First, we formulate URL-masked SMS-to-webpage fraud judgment as a benchmark setting that separates visible evidence from reputation shortcuts. Second, we construct \textbf{FraudSMSWalker}, a bilingual benchmark with hard benign cases and controlled model-visible inputs. Third, we evaluate browser-agent and text-only protocols, showing that current web agents remain weak in benign discrimination and evidence-grounded decision making.

\section{Related Work}
\label{sec:related}

Fraud-related datasets cover a wide range of modalities and threat surfaces, including financial transactions, malicious URLs, phishing webpages, fraudulent web text, telecom fraud, and smishing messages \citep{zhu2022frauddatasetbenchmark,tang-etal-2025-chifraud,ma2025teleantifraud,wang2026safeqaq,mahmood2023smishing,altay2024phishingurl,xiang2013phishingwebpage,yang2025fraudr1,wang2025smishx}. These resources support important forms of detection, but they do not directly evaluate the shortcut-suppressed chain judgment studied here. Message-only smishing benchmarks remove webpage-side evidence, while URL- or webpage-centric phishing benchmarks often expose location and reputation cues. FraudSMSWalker differs by pairing SMS context with webpage evidence and by masking URL-level signals from the model-visible input.

Web-agent benchmarks evaluate a different family of capabilities. Datasets such as Mind2Web, WebArena, WebVoyager, WebLINX, WebWalker, WebDancer, and Tongyi DeepResearch measure navigation, retrieval, page understanding, multi-step interaction, or information seeking \citep{deng2023mind2web,zhou2024webarena,he2024webvoyager,lu2024weblinx,wu2025webwalker,wu2025webdancer,tongyi2025deepresearch}. These tasks are useful for assessing whether agents can operate on the web, but their success criteria are usually task completion or answer correctness. FraudSMSWalker instead uses browser interaction to expose evidence for a safety judgment, with special emphasis on benign cases that resemble scam flows.

Our evaluation also relates to LLM-as-Judge and process-level assessment \citep{zheng2023judging,gu2024survey,shi2025judging,ratingroulette2025}. Judge-based evaluation can be sensitive to prompt design, positional bias, and self-inconsistency, and agent traces may contain reasoning that is not faithful to the executed behavior \citep{khalifa2026gaming}. We therefore do not use a judge to assign fraud labels. The judge is restricted to an evidential-support audit: given the masked prompt, browser observations, tool trajectory, and final answer, it checks whether the stated conclusion is supported by evidence available to the agent.

\begin{figure*}[t]
  \centering
  \includegraphics[width=0.75\textwidth]{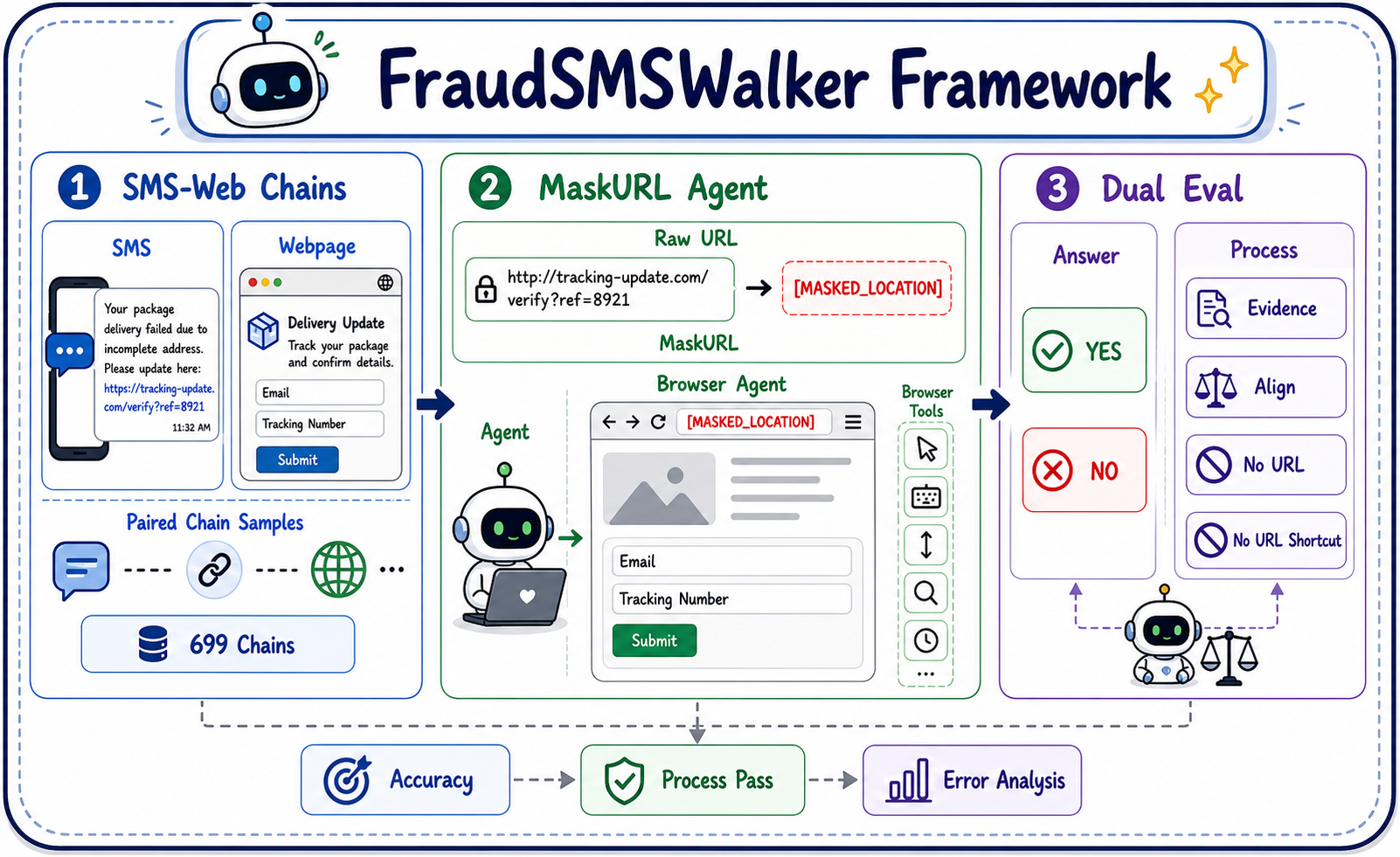}
  \caption{Overview of FraudSMSWalker. SMS-to-webpage chains are converted into URL-masked benchmark cases; an agent inspects rendered webpage evidence through controlled browser tools; evaluation checks both the final fraud decision and the evidential support of the interaction trajectory.}
  \label{fig:framework}
\end{figure*}

\section{FraudSMSWalker Benchmark}
\label{sec:benchmark}

FraudSMSWalker is designed to evaluate fraud judgment over an SMS-to-webpage chain while suppressing URL-level shortcuts. Each instance presents a message context and webpage-side evidence, but withholds raw location and reputation signals from the model-visible input. The benchmark is controlled by construction: it is intended to isolate evidence use and false-positive control, rather than to reproduce the natural distribution of SMS traffic.

\subsection{Task Definition}
\label{sec:task-definition}

Each instance pairs an SMS context with webpage evidence. The model must decide whether the combined service flow is fraudulent, rather than whether either component appears suspicious in isolation. Formally,
\begin{equation}
  x_i = (s_i, w_i), \qquad y_i \in \{0,1\},
\end{equation}
where $s_i$ is the SMS context, $w_i$ is the webpage-side evidence exposed under the protocol, and $y_i \!=\! 1$ and $y_i \!=\! 0$ denote fraudulent and benign chains. The output is standardized as \texttt{ANSWER: YES} for fraudulent chains and \texttt{ANSWER: NO} for benign chains.

Three design principles guide the benchmark. First, the evaluation unit is the chain formed by the message, the linked page, and the user action requested by the page. Second, the model-visible interface withholds direct location-bearing signals, reducing reliance on domain memorization, URL patterns, or reputation metadata. Third, the final decision should be supported by evidence that the model can observe under the protocol.

\subsection{Dataset Construction}
\label{sec:data-construction}

FraudSMSWalker contains 699 SMS-to-webpage chains: 332 fraudulent and 367 benign, with 381 Chinese and 318 English samples. Each internal record stores a stable identifier, binary label, language tag, SMS text, webpage title, visible text, form-related signals, sender brand, sender category, content topic, and a private audit reference. The ten sender categories denote service scenarios and support stratified analysis without serving as fraud-type labels.

Construction starts from webpage-side evidence. A security search system retrieves candidate pages from anti-fraud scenarios, including both phishing-like pages and benign pages from visually or semantically similar contexts. For each page, we retain a sanitized title, visible text, and form signals such as password, phone, email, text-input, form, or form-like elements. On the SMS side, each case is anchored by a real-world SMS message or SMS report associated with the corresponding service claim. For safe release, we redact or normalize sensitive fields, replace the raw URL with a fixed placeholder, and preserve the service claim and requested user action needed for chain-level judgment. The real URL and source metadata are retained only for private audit and controlled page access.

This pipeline yields a controlled and release-safe benchmark rather than a verbatim raw SMS corpus. The SMS side preserves real-world service claims after redaction or normalization, while the webpage evidence anchors the risk assessment. This design improves auditability and allows us to suppress direct URL shortcuts, but it also means the benchmark should be interpreted as a diagnostic evaluation of SMS-to-webpage chain judgment rather than as a natural-traffic prevalence estimate. Appendix~\ref{app:data-distributions} reports the sender-category and content-topic distributions.

\subsection{Labeling and Quality Control}
\label{sec:labeling-qc}

Labels follow a chain-level adjudication standard.To verify label reliability and evidence retention, we conduct a stratified human audit on a random subset of cases covering both labels, both languages, and all ten sender categories. Two annotators independently inspect the URL-masked SMS context, sanitized webpage evidence, form signals, and private audit notes, and assign chain-level fraud/benign labels according to the same adjudication standard. Disagreements are resolved by discussion with a third reviewer. We report human-human agreement, adjudicated label consistency, and the proportion of audited cases judged to retain sufficient non-reputation evidence after URL masking.

\begin{table}[t]
\centering
\small
\begin{tabular}{lr}
\toprule
Audit item & Value \\
\midrule
Audited cases & 140 \\
Human-human agreement & 96\% \\
Cohen's $\kappa$ & 0.92 \\
Agreement with released labels after adjudication & 98\% \\
Cases retaining non-reputation evidence & 96\% \\
Ambiguous or disputed cases & 2\% \\
\bottomrule
\end{tabular}
\caption{Human audit of label reliability and evidence retention under URL masking.}
\label{tab:human-label-audit}
\end{table}

A chain is fraudulent only when the SMS context and webpage interaction jointly support a deceptive flow, such as credential harvesting, abnormal payment collection, identity disclosure, refund or reward bait, transfer inducement, or other social-engineering patterns. Isolated interface cues are insufficient: a login form, payment widget, phone-number field, verification-code field, or third-party-looking page does not by itself establish fraud. This standard creates hard benign cases in which the requested action is plausible under the service context despite sharing surface forms with scam pages.

Quality control focuses on four risks. We check that the public SMS contains only the URL placeholder, that required schema fields are present, that the SMS and webpage evidence form a coherent service scenario, and that private audit metadata remains available for source verification and drift analysis. Scenario metadata such as \texttt{sender\_category} and \texttt{content\_topic} is used only for stratified analysis, not as a risk label.

\subsection{Release Boundary and Access Modes}
\label{sec:release-boundary}

FraudSMSWalker separates model-visible task evidence from private audit materials. Models see only the SMS context and sanitized webpage evidence. Stable identifiers, bookkeeping fields, and labels may be used for release and local scoring, but they are not included in model prompts. Raw URLs, hosts, domains, IP addresses, redirect chains, and reputation metadata remain private because exposing them would reintroduce the shortcut the benchmark is designed to suppress.

The benchmark supports two webpage access modes. \textbf{Sanitized-snapshot} mode builds a local HTML snapshot from sanitized title, visible text, and form signals, and is the most reproducible setting. \textbf{Live-page} mode visits the real webpage from backend metadata while keeping all model-visible observations URL-masked; it is useful for diagnosing drift, access failures, and current page state.

\subsection{Output and Scoring Interface}
\label{sec:evaluation-protocol}

The benchmark standardizes the model output and separates label correctness from evidence support. Evaluation conditions and model comparisons are described in Section~\ref{sec:evaluation}.

\paragraph{Binary correctness.}
The final response must end with exactly one parseable line: \texttt{ANSWER: YES} for fraudulent chains or \texttt{ANSWER: NO} for benign chains. Let $r_i$ be the final response for sample $i$ and $p(\cdot)$ be the parser that maps \texttt{YES} to $1$, \texttt{NO} to $0$, and malformed responses to $\varnothing$:
\begin{equation}
  z_i = p(r_i), \quad z_i \in \{0,1,\varnothing\}.
\end{equation}
We report accuracy, invalid rate, and class-wise precision and recall:
\begin{equation}
  \mathrm{Acc} = \tfrac{1}{N}\sum_{i=1}^{N}\mathbb{I}[z_i = y_i], \quad
  \mathrm{Inv} = \tfrac{1}{N}\sum_{i=1}^{N}\mathbb{I}[z_i = \varnothing].
\end{equation}
Benign recall is a central metric rather than an auxiliary one. In this benchmark, many legitimate service flows contain login, payment, or verification elements, so a useful model must avoid treating those interface cues as sufficient evidence of fraud.

\paragraph{Evidence support.}
Each agentic run produces a masked interaction trajectory
\begin{equation}
  \tau_i = \{(o_{i,t}, a_{i,t})\}_{t=1}^{T_i},
\end{equation}
where $o_{i,t}$ is the masked browser observation at step $t$, $a_{i,t}$ is the corresponding agent action, and $T_i$ is the trajectory length. Binary correctness alone cannot distinguish a well-supported decision from a lucky or unsupported one. We therefore add an LLM-as-Judge audit~\citep{zheng2023judging,gu2024survey}. The judge reads the masked prompt, $\tau_i$, browser-returned evidence, and final answer, but not the gold label. It does not relabel the sample; it only checks whether the stated conclusion is supported by evidence observed by the agent.

The judge returns structured fields for evidence sufficiency $E_i$, conclusion alignment $A_i$, hallucination severity $H_i$, and forbidden URL-style reasoning $U_i$. We summarize them with a gated support indicator:
\begin{equation}
  P_i = \mathbb{I}\big[E_i^{+} \land A_i^{+} \land H_i^{+} \land U_i^{-}\big],
\end{equation}
where superscripts indicate whether each acceptance condition is satisfied under a fixed rubric and deterministic thresholding policy applied uniformly to all evaluated models. The support indicator is diagnostic: it measures whether a decision is justified by the observed trace, not whether the decision matches the gold label. Binary correctness and evidence support are therefore reported separately.

We further validate the process judge with a human audit. Human reviewers independently score a stratified subset of agent trajectories using the same rubric for evidence sufficiency, conclusion alignment, hallucination severity, and forbidden URL-style reasoning. We then compare the binary support decision and key failure tags between the judge and human reviewers. This audit is used to ensure that the judge is treated as a scalable evidence-alignment auditor, not as an unverified substitute for human evaluation.

\begin{table}[t]
\centering
\small
\begin{tabular}{lr}
\toprule
Judge validation item & Value \\
\midrule
Audited trajectories & 315 \\
Human-judge agreement on support pass/fail & 91\% \\
Cohen's $\kappa$ for support pass/fail & 0.82 \\
Agreement on unsupported-YES tag & 93\% \\
Agreement on forbidden URL-style reasoning & 95\% \\
\bottomrule
\end{tabular}
\caption{Human validation of the LLM-as-Judge evidence-support audit. The judge is evaluated as an evidence-alignment auditor, not as a source of gold fraud labels.}
\label{tab:judge-human-validation}
\end{table}

\section{Evaluation}
\label{sec:evaluation}

We evaluate FraudSMSWalker along three dimensions: binary correctness, benign recall, and evidence support. The experiments vary two factors that are central to the benchmark design: whether the model observes webpage evidence through a browser or as sanitized text, and whether URL cues are visible. Table~\ref{tab:conditions} summarizes these conditions. The main benchmark setting is \texttt{agent+maskurl}; the other settings are used for diagnostic ablations.

\begin{table}[t]
  \centering
  \footnotesize
  \setlength{\tabcolsep}{5pt}
  \renewcommand{\arraystretch}{1.08}
  \begin{tabular}{@{}lccc@{}}
    \toprule
    Condition & Evidence & URL & Role \\
    \midrule
    \texttt{agent+maskurl} & Browser & \xmark & Main \\
    \texttt{agent+url} & Browser & \cmark & URL ablation \\
    \texttt{text+maskurl} & Text & \xmark & Browser ablation \\
    \texttt{text+url} & Text & \cmark & Text ablation \\
    \bottomrule
  \end{tabular}
  \caption{Ablation design for separating evidence access from URL visibility.}
  \label{tab:conditions}
\end{table}

  \begin{table*}[t]
    \centering
    \small
    \begin{tabular}{lrrrr}
      \toprule
      Agent & Accuracy & Fraud Recall & Benign Recall & Invalid \\
      \midrule
      \textit{Always Fraud (baseline)}
        & 47.50 & \best{100.00} & \bad{0.00} & 0.00 \\
      \textit{Always Benign (baseline)}
        & \best{52.50} & \bad{0.00} & \best{100.00} & 0.00 \\
      \midrule
      Qwen3.6-Plus
        & \second{\textbf{50.93}} & 84.04 & 20.98 & 0.00 \\
      Doubao-Seed-2.0-pro
        & 49.50 & \second{89.76} & \bad{13.08} & 0.00 \\
      Kimi-K2.6
        & 49.36 & \best{90.13} & \bad{17.27} & \bad{3.72} \\
      GLM-5-Turbo
        & 47.93 & 79.22 & 19.62 & 0.00 \\
      MiniMax-2.7
        & 47.50 & 69.18 & \second{28.14} & 0.14 \\
      DeepSeek-V4-Pro
        & 46.78 & 84.34 & \bad{12.81} & 0.00 \\
      OpenAI GPT-5.5
        & 46.78 & 65.36 & \second{29.97} & 0.00 \\
      Gemini 3.1 Flash Lite
        & 46.35 & 64.16 & \best{30.25} & 0.00 \\
      Claude Sonnet 4.6 Thinking
        & \bad{42.06} & 65.36 & 20.98 & 0.00 \\
      \bottomrule
    \end{tabular}
    \caption{Binary results on the 699-case agent+realpage+maskurl evaluation. Best and second-best model values are
    highlighted in blue; red marks especially weak benign recall, high invalid rate, or the lowest model accuracy.
    Baselines show that overall accuracy alone can be misleading under class imbalance and positive-prediction
    bias.}
    \label{tab:main_binary_results}
  \end{table*}

  \label{tab:main-results}

\subsection{Experimental Setup}

We evaluate the full 699-case benchmark to answer three questions: how accurately current web agents classify URL-masked chains, whether their decisions preserve benign recall, and whether their conclusions are supported by observed evidence. The main results use the live diagnostic instantiation of \texttt{agent+maskurl}, denoted \texttt{agent+realpage+maskurl}: the backend harness accesses the live webpage, while the model receives only URL-masked browser observations through the controlled tool. This setting exposes agents to realistic page behavior while keeping location-bearing cues hidden from the model.

The evaluated set contains 332 fraudulent and 367 benign chains, 381 Chinese and 318 English samples, and ten sender categories and webpage topics. We evaluate nine web agents: Qwen3.6-Plus, OpenAI GPT-5.5, Doubao-Seed-2.0-pro, Kimi-K2.6, GLM-5-Turbo, MiniMax-2.7, DeepSeek-V4-Pro, Gemini 3.1 Flash Lite, and Claude Sonnet 4.6 Thinking. Final responses without a parseable \texttt{ANSWER} line are counted as incorrect for accuracy and reported as invalid outputs. Class-wise precision, recall, and confusion-matrix counts are computed over parseable outputs.

All evidence-support analyses use the same judge rubric. The judge sees the masked prompt, tool trajectory, browser evidence, and final answer, but not the gold label. It checks support for the conclusion rather than relabeling the sample. Across 6{,}291 agent-case pairs, 6{,}289 judge records are valid; the two invalid records are excluded only from the affected support-rate denominators.

\paragraph{Label-prior baselines.}
We include two label-prior baselines that ignore SMS and webpage content: \textit{Always Fraud}, which predicts fraud for every case, and \textit{Always Benign}, which predicts benign for every case. Their accuracies are 47.50\% and 52.50\%, respectively. These baselines are not competitive systems; they calibrate the near-balanced label distribution and make the fraud-versus-benign recall trade-off explicit.

\subsection{Main Binary Results}
\label{sec:binary-results}

Table~\ref{tab:main-results} reports the main URL-masked browser-agent results. The most consistent pattern is a large separation between fraud recall and benign recall: all nine agents recover many fraudulent chains, but few preserve benign decisions. Fraud recall ranges from 64.16\% to 90.13\%, whereas benign recall ranges from 12.81\% to 30.25\%. Qwen3.6-Plus attains the highest accuracy at 50.93\%, while Claude Sonnet 4.6 Thinking is lowest at 42.06\%. Invalid outputs are rare, appearing only for Kimi-K2.6 (3.72\%) and MiniMax-2.7 (0.14\%).

This pattern clarifies what the benchmark is measuring. The main bottleneck is benign recognition, not fraud sensitivity. Kimi-K2.6 and Doubao-Seed-2.0-pro post the highest fraud recall yet among the lowest benign recall (17.27\% and 13.08\%). Gemini and OpenAI GPT-5.5 recover somewhat more benign cases, but still remain below 47\% accuracy. Once direct location and reputation cues are hidden, the dominant observed error is over-predicting fraud on legitimate service flows. We interpret this as a model-side failure surface rather than evidence absence in the benchmark: after URL and reputation cues are hidden, agents over-weight generic risk cues such as login forms, payment widgets, identity fields, page unavailability, or vague unofficialness, while under-weighting service-flow consistency and benign contextual cues that remain observable in the SMS--webpage chain. Thus, high fraud recall alone is not a reliable indicator of robust fraud judgment in this setting; a robust agent must also distinguish deceptive flows from legitimate service flows that share similar interface elements.Importantly, the near-baseline accuracy does not imply that URL-masked cases are intrinsically indistinguishable. The benchmark retains non-reputation evidence, including the SMS-side service claim, webpage title and visible text, form-related signals, requested user actions, and cross-channel consistency between the message and the page. In a stratified human audit, reviewers can often identify discriminative cues that are not location-bearing: benign chains typically preserve service-flow consistency and lack off-script collection or inducement, whereas fraudulent chains often contain mismatches between the SMS claim and the webpage request, excessive identity or payment collection, refund or reward bait, abnormal urgency, or other deception cues. The dominant failure is therefore not the absence of evidence after masking, but the inability of current agents to calibrate these subtle chain-level cues.

\begin{figure}[t]
  \centering
  \includegraphics[width=\columnwidth]{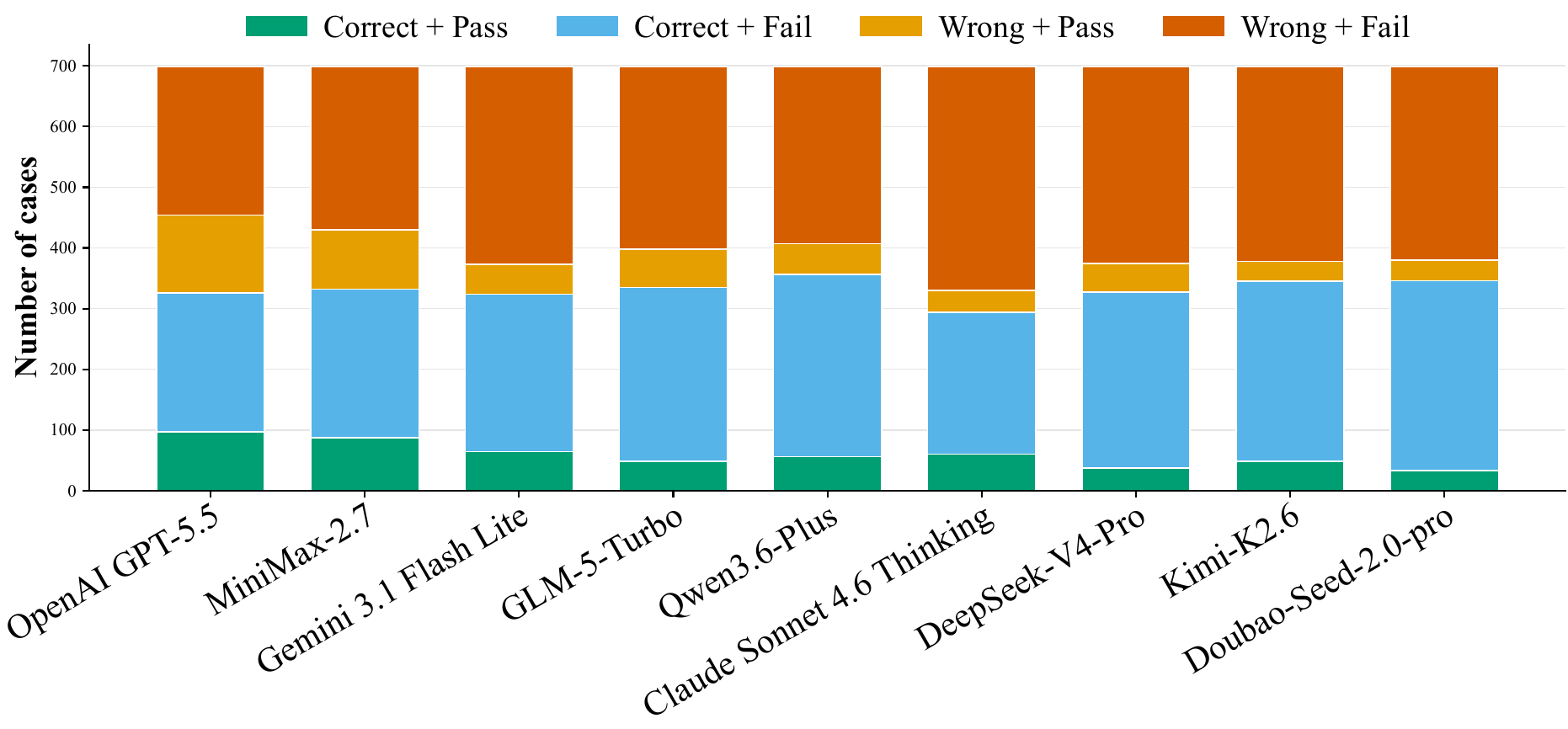}
  \caption{Counts by correctness and evidence-support status. For every agent,
  the correct-unsupported group is much larger than the correct-supported group,
  indicating that many correct labels are not well supported by the observed trajectory.}
  \label{fig:quadrants}
\end{figure}

\subsection{Process Quality Results}
\label{sec:process-results}

Binary accuracy alone hides an important distinction: an agent may output the right label while relying on weak or invented evidence. We therefore evaluate whether the final answer is supported by the masked browser trajectory. Table~\ref{tab:process-results} reports these evidence-support diagnostics. Support rates span 9.59\%--32.19\%, with OpenAI GPT-5.5 highest at 32.19\%. Even this best result means that more than two thirds of cases fail the support audit.

Unsupported positive predictions are the most stable failure mode. Eight of nine agents exceed 63\% unsupported-YES, peaking at 85.26\% for Doubao-Seed-2.0-pro and 80.11\% for DeepSeek-V4-Pro. Claude Sonnet 4.6 Thinking is the sole exception at 0.00\%, but this does not imply stronger grounding: it still has low support rate (13.73\%) and high forbidden URL-signal use (49.93\%), so its failures are captured by other audit gates. The observed fraud recall is often accompanied by overclaiming from weak signals, such as login forms, payment fields, inaccessible pages, or vague unofficialness, rather than careful evidence integration.

The audit also exposes a second failure mode: reputation-like reasoning persists even when URL cues are hidden. Gemini and OpenAI GPT-5.5 stay lower at 28.76\% and 25.75\%, while most agents fall between 42\% and 55\%. Masking the URL suppresses direct domain-reputation lookup but does not eliminate unsupported claims about official domains, low-trust sites, or URL credibility. This behavior matters because the benchmark deliberately withholds such evidence from the model-visible trace.

\subsection{Correctness versus Evidence Support}
\label{sec:quadrants}

We further cross-tabulate binary correctness with evidence support. Figure~\ref{fig:quadrants} shows that, for every agent, correct-but-unsupported predictions far exceed correct-and-supported ones: Qwen3.6-Plus has 300 versus 56, and Doubao-Seed-2.0-pro has 313 versus 33. Thus, many correct labels arise without sufficient trajectory evidence. Evidence support therefore serves as a complementary diagnostic rather than a replacement for binary accuracy: it reveals when high recall reflects an aggressive fraud bias rather than grounded chain-level reasoning.

\begin{figure}[t]
  \centering
  \includegraphics[width=\columnwidth]{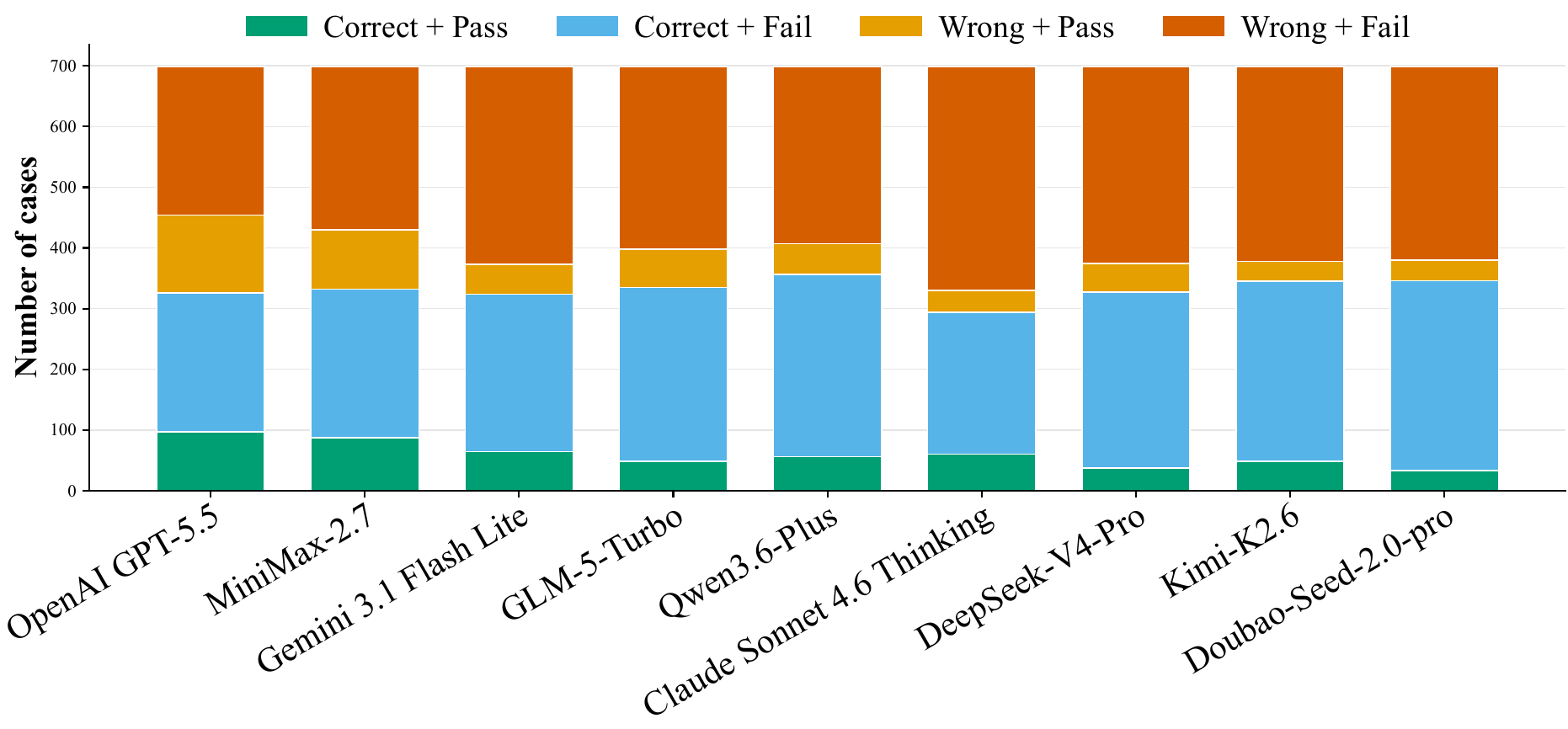}
  \caption{Counts by correctness and evidence-support status. Correct-but-unsupported predictions dominate across agents.}
  \label{fig:quadrants}
\end{figure}

\subsection{Stratified Results and Error Analysis}
\label{sec:error-analysis}

Errors concentrate in legitimate service flows that resemble scam lures. Government, logistics, e-commerce, banking, travel, payment, and telecom pages often contain normal login, verification, order-query, payment, or account-management elements. Agents frequently treat these interface elements as sufficient fraud evidence, producing the low benign recall in Table~\ref{tab:main-results}. The dominant false-positive mechanisms are \textit{interface overclaiming} on routine forms or payment widgets, \textit{availability overclaiming} on page-load failures or redirects, and \textit{ungrounded reasoning} that invents page content or unsupported risk claims. False negatives mainly reflect missed credential harvesting or live-page drift, which is why FraudSMSWalker retains masked traces and private audit metadata for diagnosis.

\begin{figure*}[t]
  \centering
  \includegraphics[width=0.90\textwidth]{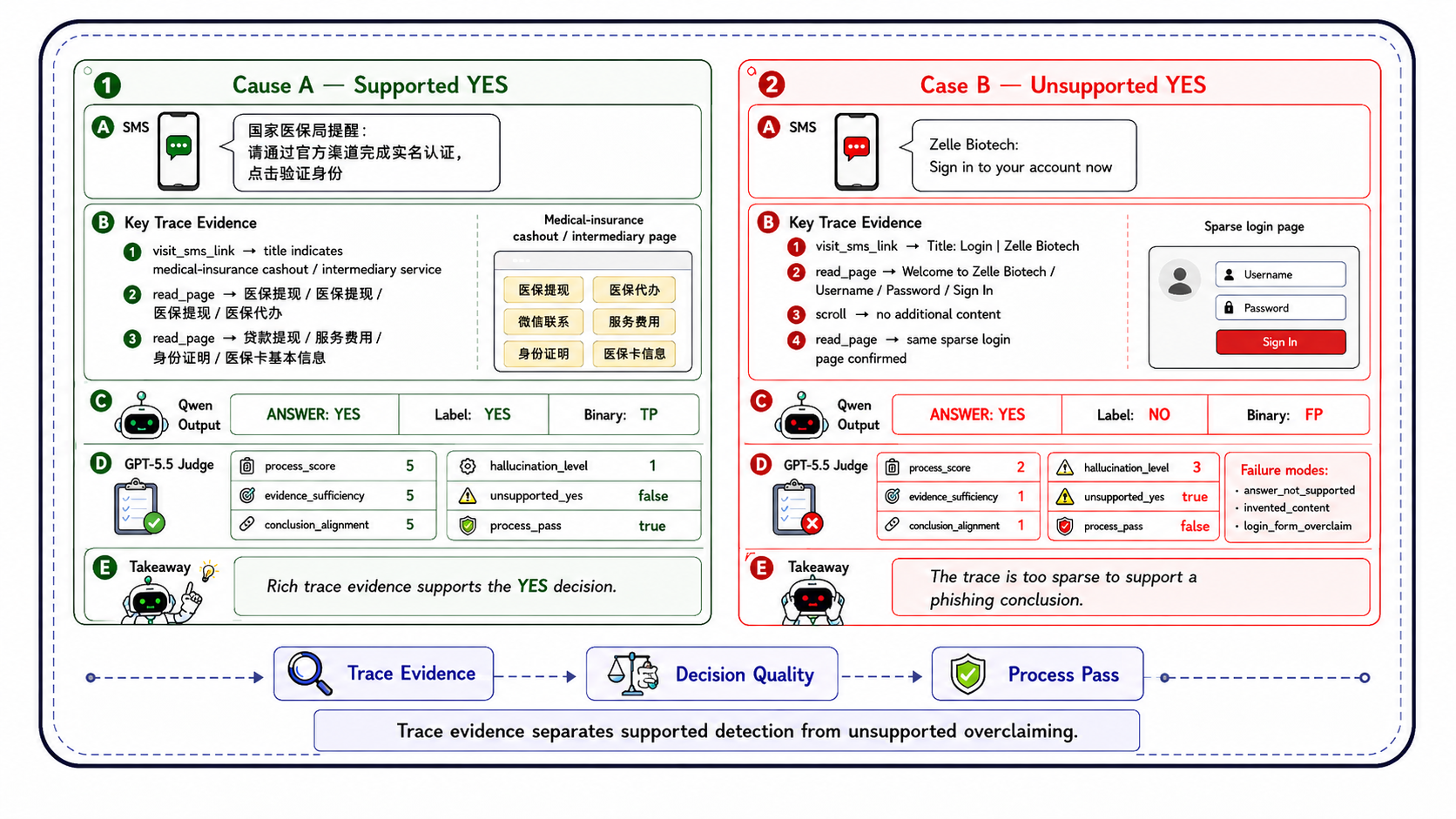}
  \caption{\textbf{Case study of trace-based judgment for phishing detection.} Trace evidence separates supported detection from unsupported overclaiming. A supported YES decision is grounded in rich page-level evidence, while an unsupported YES is flagged when the trace is too sparse to justify the phishing conclusion.}
  \label{fig:case_study}
\end{figure*}

\subsection{Trace Analysis}

In Figure~\ref{fig:case_study},the supported case (left) shows that the model’s YES decision is grounded in rich behavioral trace evidence, including SMS-link context and multiple page observations consistent with medical-insurance cashout/intermediary services, leading to a true positive and a passing process score. In contrast, the unsupported case (right) shows a YES decision made by over-interpreting a routine login interface. The trace contains a generic login page that is consistent with a benign service flow and does not show fraud-specific cues such as off-script payment, reward bait, credential harvesting beyond the stated service context, or mismatch between the SMS claim and the requested action. In the right figure, the model jumps to a conclusion without sufficient evidence, resulting in a false positive where a benign website is misclassified as fraudulent. This case is consequently flagged as a false positive and labeled by the judge as an unsupported overclaim. This comparison illustrates how trace evidence helps distinguish supported chain-level detection from unsupported interface overclaiming. The key issue is not that masked evidence is absent, but that agents often fail to use negative and contextual evidence, such as service-flow consistency and the absence of fraud-specific requested actions, when deciding whether a routine-looking page should remain benign.

\subsection{Ablation Study}
\label{sec:ablation}

To separate the effects of browser interaction and URL visibility, we run a 2-by-2 diagnostic ablation with Qwen3.6-Plus under the \textbf{sanitized-snapshot} protocol, in which both agent and text-only cells consume the same sanitized webpage evidence (page title, visible text, and form signals) drawn from a fixed local HTML snapshot. This isolates URL visibility and tool-use as the only varying factors and removes confounds from live-page drift and network behavior. Because the snapshot protocol differs from the live-page setting of Table~\ref{tab:main-results}, ablation accuracies are not directly comparable to the live-page leaderboard and are reported only for within-grid directional analysis.

\begin{table}[t]
  \centering
  \small
  \setlength{\tabcolsep}{3pt}
  \resizebox{\columnwidth}{!}{%
  \begin{tabular}{lrrrrr}
    \toprule
    Setting & Acc. & Fraud Recall & Benign Recall & FP & FN \\
    \midrule
    \texttt{snapshot-agent+url}     & 49.79 & 75.90 & 26.16 & 271 & 80  \\
    \texttt{snapshot-agent+maskurl} & 57.80 & 29.82 & 83.11 & 62  & 233 \\
    \texttt{text+url}               & 52.22 & 46.99 & 56.95 & 158 & 176 \\
    \texttt{text+maskurl}           & 57.22 & 35.84 & 76.57 & 86  & 213 \\
    \bottomrule
  \end{tabular}}
  \caption{Qwen3.6-Plus diagnostic ablation under the
  sanitized-snapshot/textual protocol. All cells use the same 699
  samples with zero invalid outputs. Scores are not directly comparable
  to the live-page leaderboard in Table~\ref{tab:main-results}.}
  \label{tab:ablation}
\end{table}

The ablation shows that URL visibility mainly shifts the fraud--benign operating point. In the agent setting, revealing URLs raises fraud recall from 29.82\% to 75.90\%, but reduces benign recall from 83.11\% to 26.16\%. Trace inspection suggests that URL-visible agents often treat non-official domains or IP-based hosts as fraud cues, creating many false positives on benign chains. With URLs masked, the model must compare the SMS claim with sanitized page evidence, restoring benign recall and improving accuracy in this snapshot setting. The text-only comparison follows the same direction, while \texttt{snapshot-agent+maskurl} and \texttt{text+maskurl} remain close in accuracy (57.80\% vs.\ 57.22\%), suggesting that browser interaction adds inspectable trajectories but limited binary-decision gain once URL shortcuts are removed.

\section{Conclusion}

FraudSMSWalker is a URL-masked benchmark for SMS-to-webpage fraud judgment. It is designed to suppress direct URL and reputation shortcuts and to stress hard benign cases that require false-positive control. Across nine web agents, the dominant pattern is clear: models often detect suspicious signals, but they struggle to recognize benign service flows and to support positive predictions with observed evidence. Ablation study shows, for Qwen3.6-Plus in the snapshot setting, that URL visibility changes the fraud-versus-benign operating point more than overall accuracy. FraudSMSWalker therefore motivates evaluation protocols that report not only final labels, but also benign recall and evidence support under controlled model-visible evidence.

\section*{Limitations}

FraudSMSWalker is a controlled benchmark rather than a raw sample of naturally observed SMS traffic. Although the SMS side is anchored by real-world SMS messages or reports, sensitive fields are redacted or normalized for safe release, which may reduce linguistic diversity compared with verbatim SMS corpora. Live-page evaluation introduces drift as pages disappear, redirect, or change. Residual stylistic cues may remain after removing URLs, hosts, domains, IPs, and reputation metadata. Finally, the support audit is judge-based, so we validate it with a stratified human audit rather than treating it as a substitute for human security review.


\bibliography{custom}

\clearpage
\appendix

\section{Additional Dataset Statistics}
\label{app:data-distributions}

This appendix reports the dataset distributions used for stratified
analysis. Sender categories describe SMS-side service scenarios rather
than fraud types, while webpage topics describe visible interaction
patterns on the linked webpages.

\begin{figure}[H]
  \centering
  \includegraphics[width=\columnwidth]{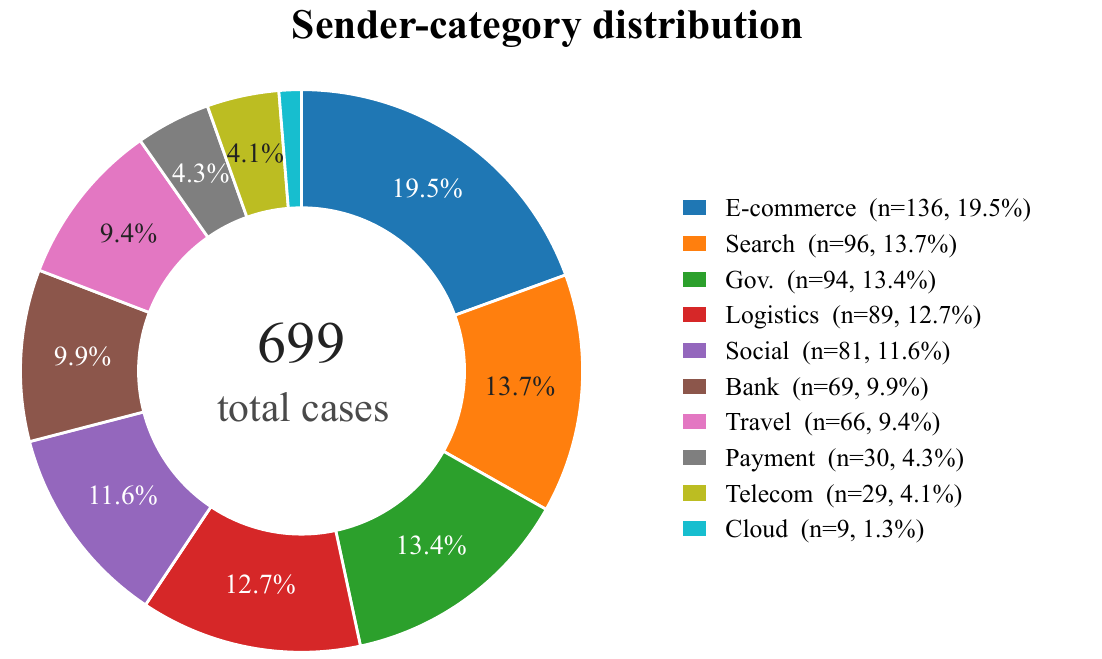}
  \caption{Sender-category distribution. Categories denote SMS-side service scenarios, not fraud types.}
  \label{fig:app-sender-dist}
\end{figure}

\begin{figure}[H]
  \centering
  \includegraphics[width=\columnwidth]{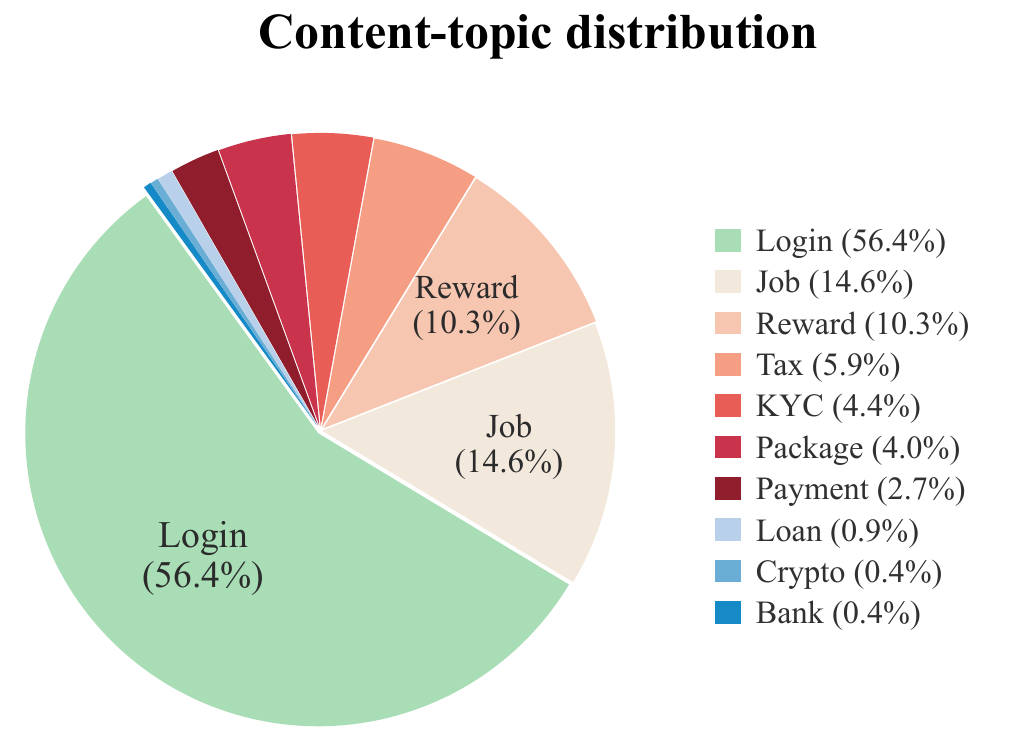}
  \caption{Webpage content-topic distribution. Topics describe visible interaction patterns for stratified error analysis.}
  \label{fig:app-topic-dist}
\end{figure}

\end{document}